\documentclass[letterpaper]{article} 
\usepackage{aaai2027}  
\usepackage[hyphens]{url}  
\usepackage{graphicx} 
\usepackage{booktabs}
\usepackage{natbib}  
\usepackage{caption} 
\usepackage{amsmath}
\usepackage{amssymb}
\usepackage{amsthm}
\usepackage{booktabs}
\usepackage{adjustbox}
\usepackage{multirow}
\usepackage{graphicx}
\newcommand{\mlogo}[2][3ex]{%
  \raisebox{-0.30\height}{\includegraphics[height=#1]{figs/logo/#2}}\,}

\usepackage{tabularx}
\usepackage{caption}
\usepackage{amsmath, amssymb, amsthm}

\newtheorem{definition}{Definition}

\usepackage{amsmath,amssymb}
\usepackage[ruled,vlined,linesnumbered]{algorithm2e}
\usepackage{xcolor}

\SetCommentSty{textnormal}
\DontPrintSemicolon
\usepackage{bm}
\usepackage{comment}
\title{Functional Depth Is Learned:\\Isolating the Computation That Layers Produce Only Together}

\author{
    JaeSeong Kim\textsuperscript{\rm 1}\thanks{Work done during an internship at MLP Lab, KAIST.},
    KyungTae Lim\textsuperscript{\rm 2}\thanks{Corresponding author.}
}
\affiliations{
    \textsuperscript{\rm 1}Semyung University, Jecheon-si, Republic of Korea\\
    \textsuperscript{\rm 2}Korea Advanced Institute of Science \& Technology\\
    kjsqp1010@semyung.ac.kr, ktlim@kaist.ac.kr
}

\begin{document}

\maketitle

\begin{abstract}

Transformer performance scales with depth, but layer count merely counts the opportunities for sequential computation: layers could still behave as nearly independent refinements that barely depend on one another. We study how transformers develop functional depth during pretraining, introducing an exact inclusion–exclusion decomposition that isolates the connected component—the computation that appears only when a specified set of layers acts together under identity ablation. Tracking this quantity across public checkpoints of Pythia, OLMo, and Amber yields three findings. First, write-relative connected coordination strengthens from its initialization or earliest-checkpoint baseline as pretraining proceeds, across all measured interaction orders. Second, these connected components are behaviorally consequential under direct intervention: deleting them damages perplexity more than deleting a norm-matched random direction at the same site in all 35 measured model–order cells. Third, pairwise representation similarity is consistently negatively correlated with pairwise connected coordination across seven models, yet the two signals are not interchangeable: similarity- and coordination-based block-removal criteria often agree in Pythia but diverge on Amber. Together, these results suggest that architectural depth provides the opportunity for sequential computation, while pretraining progressively organizes cross-layer dependencies within that depth.

\end{abstract}

\section{Introduction}

Depth is one of the few reliable scaling axes of modern transformers, and recent studies have shown that stacking layers is central to improving model performance \cite{kaplan2020scaling, tay2022scale, petty2024impact}. Unfortunately, the mechanism behind what added depth actually buys remains poorly understood. A stacked architecture offers as many opportunities for sequential computation as it has layers, yet a trained model could still behave as a collection of nearly independent refinements \cite{veit2016residual, jastrzebski2018residual}, with each layer editing the residual stream \cite{elhage2021mathematical} without depending on combinations of earlier computations. Architectural depth, in other words, is only potential depth. Existing analysis tools---representation similarity \cite{raghu2017svcca, kornblith2019similarity, nguyen2021wide}, layer lenses \cite{belrose2023eliciting}, and layer-skipping and reordering studies \cite{freiberger2024layershuffle, sun2025painters}---inspect layers individually or in pairs; they can measure what each layer leaves behind, but not the computation that multiple layers produce only when present together.

In this work, we take an initial step toward understanding the dynamics by which functional depth is formed during transformer pretraining. We investigate three important research questions that remain unanswered:

\begin{itemize}
    \item[\textbf{RQ1.}] Do the layers of a trained transformer actually compute jointly across depth, or do they behave as a stack of independent refinements? And how does this cross-layer coupling structure change over the course of pretraining?
    \item[\textbf{RQ2.}] Is the computation that layers produce only together 
    behaviorally consequential under direct intervention, relative to an equal-norm perturbation at the same site?
    \item[\textbf{RQ3.}] Does cross-layer connective structure provide a complementary account of the phenomena reported by layerwise and pairwise analyses? 
\end{itemize}

\begin{figure*}[ht]
\centering
\includegraphics[width=0.95\textwidth]{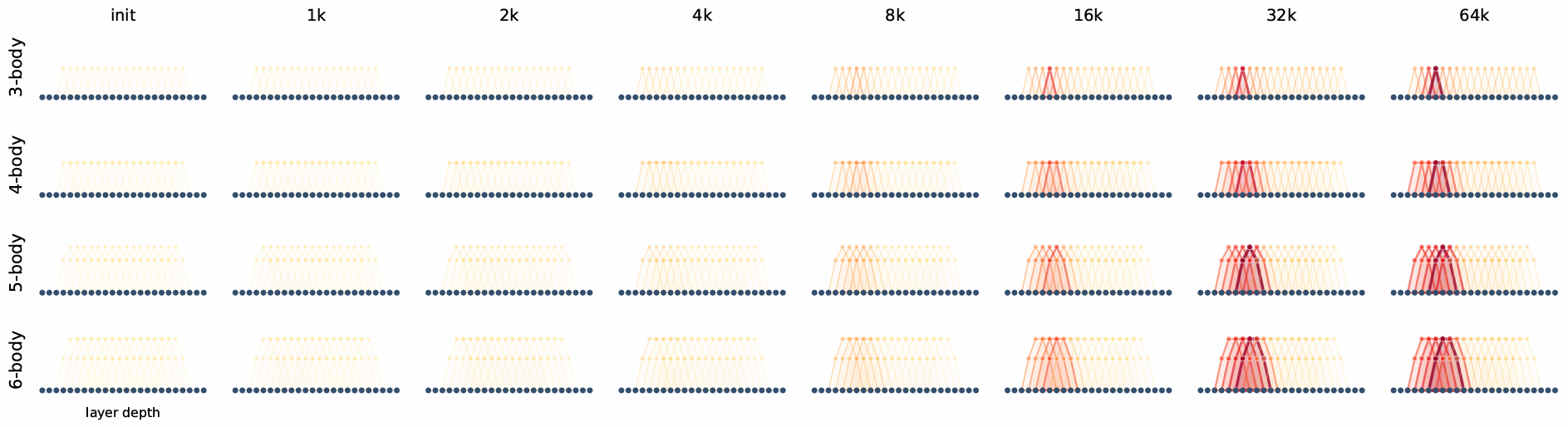}
\caption{Development of connected cross-layer coordination during pretraining
(pythia-410m). Each panel visualizes one checkpoint (columns, init--64k steps) at one
interaction order $k$ (rows, $k=3$--$6$): nodes are layers ordered by depth, and each
arc spans an adjacent window $T$ of $k$ layers, with intensity proportional to its
normalized connected component $\widehat{U}_T$, scaled by
the order-level ratio $R_k$. 
Coordination is weak relative to later checkpoints at initialization and progressively concentrates into an early-interior band as training proceeds, across every measured order.
}
\label{fig:figure1}
\end{figure*}

Our experiments reveal several important insights into the fine-grained dynamics by which functional depth is formed during transformer pretraining.


First, we find that functional depth, operationalized by write-relative connected components, strengthens during pretraining (Figure~\ref{fig:figure1}). At initialization or at the earliest released checkpoints \cite{biderman2023pythia, groeneveld2024olmo}, normalized connected components are substantially smaller, particularly at higher interaction orders, whereas later checkpoints exhibit larger connected components across all measured orders and depths. Our normalization expresses each connected component relative to the actual write of the final participating layer, allowing us to compare interaction strength against a local write scale rather than raw residual magnitude. The resulting pattern is consistent with pretraining progressively organizing cross-layer interactions across depth. Second, the connected component is causally load-bearing rather than an algebraic remainder. Deleting it damages perplexity more than deleting a norm-matched random direction of the same size at the same site, in all 35 of our measured model--order cells, with the gap exceeding two orders of magnitude in the largest models. Third, representation similarity is systematically associated with weak cross-layer coordination, while the two quantities remain distinct. 
Across seven models differing in scale and training data, pairwise similarity is consistently negatively correlated with the normalized connected component, so layer pairs with more similar representations tend to exhibit weaker pairwise connected computation. 

Taken together, to the best of our knowledge, this work is among the first attempts to isolate cross-layer joint computation in transformers with an exact estimand and to trace its development. Layer count specifies only the possibility of depth; whether that possibility becomes an organized computation is decided by pretraining. By treating functional depth as an empirical property of a trained computation rather than an architectural hyperparameter, we hope to help the community build a holistic understanding of deep models and make better-informed decisions about pruning, compression, and scaling.

\section{Related Work}
\label{sec:related}
Most analyses of transformer depth build on the residual-stream view of each
block as an identity map plus a per-layer update~\cite{elhage2021mathematical},
and inspect one or two layers at a time: decoding the state at each
layer~\cite{nostalgebraist2020logitlens, belrose2023eliciting}, comparing pairs
of representations~\cite{raghu2017svcca, kornblith2019similarity,
nguyen2021wide}, or scoring single blocks for skipping, reordering, or
pruning~\cite{freiberger2024layershuffle, sun2025painters,
gromov2024unreasonable, men2024shortgpt}. Causal-intervention methods likewise
localize behavior by patching activations at chosen
sites~\cite{meng2022locating, wang2023interpretability,
goldowskydill2023localizing, chan2022causal, zhang2024best}; the cross-layer
attribution methods of \citet{lindsey2024crosscoders},
\citet{dunefsky2024transcoders}, and \citet{lindsey2025biology} are closest in
spirit, but decompose the \emph{representation} into features spanning layers.
We instead decompose the \emph{function}: whether one layer's write changes how
later layers compute, including effects requiring three or more layers jointly,
with the residual identity as the null---the only value under which $U_T$
vanishes exactly when a participating layer is removed
(Section~\ref{sec:cka} contrasts the resulting block-removal rankings with
similarity-based ones).
Residual networks have been read as ensembles of shallow
paths~\cite{veit2016residual} or as iterative
inference~\cite{jastrzebski2018residual}; the former enumerates subpaths,
whereas we isolate what the full path contributes beyond every proper subpath.
Formally, our estimand is the M\"obius transform, or Harsanyi dividend, of the
set function induced by layer ablation~\cite{harsanyi1963simplified,
grabisch1999axiomatic}, standard for input features and
tokens~\cite{sundararajan2020shapley, tsai2023faithshap, hao2021self}. We
differ in that the payoff is a vector-valued residual write, the players are
noncommutative maps, and the residual-identity null makes the decomposition
exact; connecting to studies of training
dynamics~\cite{nanda2023progress, olsson2022incontext,
hoogland2024developmental}, we track this quantity across pretraining, asking
not what a model represents but how many layers must act together before a
computation exists at all.

\section{Cross-Layer Coordination}
\label{sec:coordination}
The number of layers the architecture supplies must be distinguished
from the depth the trained model actually uses: if each layer edited
the residual stream independently of the computations before it, the
model would behave less like a deep computation than like a collection
of independent refinements applied in sequence.
We therefore define \emph{cross-layer coordination} as follows.
If a later layer produces the same update regardless of the outputs of
earlier layers, the layers compute independently; if the results of
earlier layers change what the later layer computes, the layers
coordinate.
Coordination is thus not several layers producing large updates, but
one layer's computation depending conditionally on the computations
that precede it.
Our goal is to isolate the component that appears only when a specified
set of layers all act together, and that is not explained by any proper
subset.

\subsection{The Connected Component as Conditional Computation}
\label{sec:connected-component}

Consider a pretrained transformer with $L$ layers, and let
$h^{(\ell)}\in\mathbb{R}^{d}$ denote the residual-stream state entering
layer $\ell$.
Each transformer block acts as
\begin{equation}
f_\ell(h)=h+g_\ell(h)
\label{eq:block}
\end{equation}
where $g_\ell:\mathbb{R}^{d}\to\mathbb{R}^{d}$ is the full update that
layer $\ell$'s attention and MLP write to the residual stream.

Consider first two layers $i<j$ acting in sequence, and expand the
forward pass directly:
\begin{equation}
\bigl(f_j\circ f_i\bigr)(h)
=\underbrace{\bigl(h+g_i(h)\bigr)}_{\text{output of layer } i}
+\;g_j\bigl(h+g_i(h)\bigr)
\label{eq:two-layer-forward}
\end{equation}
In the last term of Eq.~\eqref{eq:two-layer-forward}, layer $j$
receives as input not the original state $h$ but the state to which
layer $i$'s update has already been written: what layer $j$ computes
depends on what layer $i$ wrote.
To isolate this dependence, we split layer $j$'s update into a part
determined independently of layer $i$ and a part that is not:
\begin{equation}
\delta_{ij}(h):=g_j\bigl(h+g_i(h)\bigr)-g_j(h)
\label{eq:delta}
\end{equation}
$g_j(h)$ is the update layer $j$ produces even without layer $i$'s
participation, and $\delta_{ij}(h)$ is the extent to which layer $i$'s
write changed layer $j$'s computation.
If $\delta_{ij}(h)=0$, the two layers are merely placed in sequence and
compute independently; if $\delta_{ij}(h)\neq 0$, the two layers
coordinate.
$\delta_{ij}$ is therefore a computation that exists only when the two
layers are present together.

This quantity can be rewritten in terms of subset forward passes.
For a subset $S=\{s_1<\cdots<s_m\}\subseteq T$ of a layer set $T$,
define the result of applying only the layers in $S$, in their original
depth order, as
\begin{equation}
F^{S}(h)=\bigl(f_{s_m}\circ\cdots\circ f_{s_1}\bigr)(h),
\qquad F^{\emptyset}(h)=h
\label{eq:subset-forward}
\end{equation}
with every unselected layer replaced by the identity map.
In this notation $F^{\{i,j\}}(h)$ is Eq.~\eqref{eq:two-layer-forward},
and $F^{\{i\}}(h)=h+g_i(h)$, $F^{\{j\}}(h)=h+g_j(h)$, so
\begin{equation}
\begin{split}
U_{\{i,j\}}(h)
&:= F^{\{i,j\}}(h)-F^{\{i\}}(h)-F^{\{j\}}(h)+F^{\emptyset}(h)\\
&\;= \delta_{ij}(h)
\end{split}
\label{eq:pairwise-connected}
\end{equation}
Eq.~\eqref{eq:pairwise-connected} takes the result of applying the two
layers together, removes the effects that each layer explains on its
own, and leaves only the component that arises from their sequential
composition.
The $\delta_{ij}$ obtained by expanding the forward pass thus coincides
exactly with the result of an inclusion--exclusion operation over the
layer set, and this form generalizes directly to three or more layers.

\subsection{Higher-Order Connected Components}
\label{sec:higher-order-connected}

Extending the two-layer definition to three or more layers requires
more than inspecting the joint application of all the layers.
The full forward pass of $k$ layers contains not only the computation
that requires all of them together, but also every lower-order
contribution explained by individual layers and smaller subsets.
Isolating genuine order-$k$ coordination therefore requires removing
these lower-order components.

For three layers $T=\{1,2,3\}$, for example, the component that
requires all three is
\begin{align}
U_{\{1,2,3\}}(h)
={}&
F^{\{1,2,3\}}(h)
\nonumber\\
&-
\Bigl(
F^{\{1,2\}}(h)
+
F^{\{1,3\}}(h)
+
F^{\{2,3\}}(h)
\Bigr)
\nonumber\\
&+
\Bigl(
F^{\{1\}}(h)
+
F^{\{2\}}(h)
+
F^{\{3\}}(h)
\Bigr)
-
F^{\emptyset}(h)
\label{eq:three-way-connected}
\end{align}
We take the result of all three layers, remove the effect of each pair
of layers, add back the single-layer effects that were removed twice in
the process, and finally subtract the untouched input.
This inclusion--exclusion cancels every lower-order contribution
exactly.

We generalize to arbitrary order.
Let
\begin{equation}
T=(\ell_1<\ell_2<\cdots<\ell_k)
\end{equation}
be a set of layers with their original depth order retained.

\begin{definition}[Connected component]
\label{def:connected-component}
The connected component of a layer set $T$ is
\begin{equation}
U_T(h)
=
\sum_{S\subseteq T}
(-1)^{|T|-|S|}F^{S}(h)
\label{eq:connected-component}
\end{equation}
\end{definition}

$U_T$ is the computation that appears when all selected layers act
together but is not explained by the forward pass of any proper subset
$S\subsetneq T$.
It is thus a vector-valued expression of the coordination of the layer
set $T$: $U_T(h)\neq 0$ means that the joint participation of all
selected layers produces a change that does not reduce to the
computation of lower-order layer sets.

Distinguishing connected components of this kind that are already
present at initialization from coordination formed by training,
however, requires accounting not only for the absolute size of $U_T$
but also for the scale of the layer update in which this component
appears.
Because the magnitude of a layer's write to the residual stream can
change over the course of pretraining, an increase in
$\lVert U_T\rVert$ alone does not establish that the connected
component has become larger relative to the computation produced at
that layer.
The structure of $U_T$ itself provides a natural normalization.
Writing $\ell_k=\max T$ and $T_{<\ell_k}=T\setminus\{\ell_k\}$, and
splitting the sum in Eq.~\eqref{eq:connected-component} by whether
$\ell_k$ is included, we obtain
\begin{equation}
U_T(h)
=
\sum_{S\subseteq T_{<\ell_k}}
(-1)^{\,k-1-|S|}\,
g_{\ell_k}\!\bigl(F^{S}(h)\bigr)
\label{eq:last-layer-form}
\end{equation}

Every term in Eq.~\eqref{eq:last-layer-form} is an update produced by
the final participating layer $\ell_k$, evaluated under a different
subset-conditioned input.
Thus, $U_T$ isolates the part of layer $\ell_k$'s write associated
with the joint presence of all preceding participating layers beyond
what is explained by their proper subsets, and a natural local scale
for its magnitude is layer $\ell_k$'s actual write when all preceding
participating layers are present.
We therefore define the normalized connected component of order
$k=|T|$ as
\begin{equation}
\widehat{U}_T(h)
=
\frac{
\left\lVert U_T(h)\right\rVert
}{
\left\lVert
g_{\ell_k}\!\bigl(F^{T_{<\ell_k}}(h)\bigr)
\right\rVert
+\varepsilon
}
\label{eq:normalized-connected}
\end{equation}
where $\varepsilon>0$ is a small constant for numerical stability.
The denominator is the term in Eq.~\eqref{eq:last-layer-form}
corresponding to $S=T_{<\ell_k}$, i.e., the magnitude of the update
that layer $\ell_k$ actually writes when all preceding participating
layers are present.
$\widehat{U}_T$ therefore measures the magnitude of the order-$k$
connected component relative to the actual write of the final
participating layer.
This normalization expresses the connected component in units of the
local write scale of layer $\ell_k$, rather than in absolute residual
units.
It does not separately normalize the magnitudes of the preceding
layers' updates, whose effects enter through the subset-conditioned
inputs $F^S(h)$.

\paragraph{When $\widehat{U}_T$ is close to 0.}
Layer $\ell_k$'s update does not require the joint computation of the
preceding layers.
This covers not only the case where layer $\ell_k$ writes the same
update regardless of what the preceding layers computed, but also the
case where each preceding layer does change the update, yet the change
reduces to a sum of per-layer contributions.
In the latter case layer $\ell_k$ coordinates with the preceding layers
only individually, and such lower-order coordination cancels exactly in
the alternating sum of Eq.~\eqref{eq:connected-component}.
Either way, no order-$k$ coordination exists.

\paragraph{When $\widehat{U}_T$ is close to 1.}

The order-$k$ connected component has a magnitude comparable to the actual update written by layer $\ell_k$ when all preceding participating layers are present. Thus, the joint interaction associated with all layers in $T$ is large relative to the local write scale of the final participating layer.
Importantly, $\widehat{U}_T$ should not be interpreted as the fraction of layer $\ell_k$'s update attributable to order-$k$ coordination, because the connected and lower-order components need not be orthogonal. For the same reason, $\widehat{U}_T$ is not bounded above by 1: $U_T$ is an alternating combination of actual and counterfactual updates, and its norm can exceed that of the actual write used for normalization.


Finally, to assess how pretraining changes cross-layer coordination, we compare the normalized connected component of a trained
checkpoint to that of the same architecture at the baseline checkpoint, defining the final coordination ratio at order
$k$ as
\begin{equation}
R_k
=
\frac{
\widehat{U}^{\,\mathrm{final}}_T(h)
}{
\widehat{U}^{\,\mathrm{base}}_T(h)
}
\label{eq:coordination-ratio}
\end{equation}

where the baseline is the initialization checkpoint when available and otherwise the earliest released checkpoint. $R_k = 1$ means that the model's order-$k$ write-relative connected component remains at its baseline level, whereas $R_k > 1$ indicates strengthening relative to that baseline. We use $R_k$ to quantify how cross-layer interaction changes during pretraining relative to its order-specific starting level. Because the baseline value of $\widehat{U}_T$ decreases with interaction order, $R_k$ should be interpreted within each order rather than as an absolute comparison of interaction strength across orders.


\section{Experimental Setup}
\label{sec:setup}

\paragraph{Models.}
We use public checkpoints of Pythia~\cite{biderman2023pythia},
OLMo~\cite{groeneveld2024olmo}, and Amber~\cite{liu2023llm360}.
All three families release intermediate pretraining checkpoints, which
lets us track how coordination is organized over the course of training
within a fixed architecture.
Using families that differ in scale, depth, and training data guards
against the observations being a property of any single model family.
The models and checkpoints are summarized in Table~\ref{tab:models}.

\begin{table}[ht]
\centering
\small
\setlength{\tabcolsep}{4pt}
\begin{tabular}{@{}l l c c l@{}}
\toprule
Series & Models & $L$ & $d$ & Checkpoints \\
\midrule
\multirow{5}{*}{\mlogo{pythia.png}Pythia}
& 410M & 24 & 1024 & \multirow{5}{*}{8 (0--64k steps, 134B tok.)} \\
 & 1B   & 16 & 2048 & \\
 & 1.4B & 24 & 2048 & \\
 & 2.8B & 32 & 2560 & \\
 & 6.9B & 32 & 4096 & \\
\mlogo[2.5ex]{olmo.jpeg}OLMo & 1B & 16 & 2048 & 7 (4B--3T tok.) \\
\mlogo{amber.jpeg}Amber      & 7B & 32 & 4096 & 8 ($\leq$1.26T tok.) \\
\bottomrule
\end{tabular}
\caption{Model series and released checkpoints used in the developmental analysis.}
\label{tab:models}
\end{table}

\paragraph{Dataset.}
All measurements are performed on
WikiText-103~\cite{merity2017pointer}.
Residual-stream states over contiguous text chunks serve as the base
point $h$ for every estimand; $U_T(h)$ is computed independently at
each token position and then averaged.
The perplexity evaluations of Section~\ref{sec:coord-causal} and the
CKA~\cite{kornblith2019similarity} measurements of
Section~\ref{sec:cka} use the same corpus and the same token sample, so
coordination, similarity, and deletion damage are all compared on the
same distribution.

\paragraph{Probed tuples.}
All measurements are taken over tuples of layers adjacent in depth.
An order-$k$ tuple is one of the adjacent windows
\begin{equation}
  T = \{\ell,\, \ell+1,\, \dots,\, \ell+k-1\}
  \label{eq:window}
\end{equation}
of which a model of depth $L$ has $L-k+1$.
We restrict to adjacent windows so that the subset forward passes and
the full forward pass share the same input state and bypass no
intermediate layer, keeping the site of the deletion intervention well
defined.

\paragraph{Interventions.}
The deletions of Section~\ref{sec:coord-causal} proceed as follows.
During the full forward pass we record the base point $h$ at the input
of $\ell_1$, run the $2^{k}$ subset forward passes separately to
compute $U_T(h)$, subtract it from the residual stream immediately
after the output of block $\ell_k$, and run the remaining layers
unchanged.
Because subset composition preserves depth order, the alternating sum
of Eq.~\eqref{eq:connected-component} resolves exactly into the write
of the last participating layer, so the intervention site is determined
structurally as $\ell_k$ rather than by approximation.
The deletion thus leaves the model to continue computing from the best
subset-explainable reconstruction of its own state.
The random control subtracts, at the same site, a Gaussian direction
rescaled per token to $\lVert U_T \rVert$.
The two interventions match in norm exactly at the point of action, so
any difference in damage is attributable solely to whether the deleted
direction is the connected component.
Reported factors are post-intervention perplexity divided by
unintervened perplexity.

\section{Results}

 \subsection{Coordination Strengthens During Pretraining}
\label{sec:coord-learned}

\begin{table}[thb]
\centering
\small
\setlength{\tabcolsep}{3pt}
\begin{tabular}{@{}l l rrrrr@{}}
\toprule
\multirow{2}{*}{Model} & & \multicolumn{5}{c}{Order $k$} \\
\cmidrule(lr){3-7}
 & & $k{=}2$ & $k{=}3$ & $k{=}4$ & $k{=}5$ & $k{=}6$ \\
\midrule
\multirow{3}{*}{\raisebox{-0.30\height}{\includegraphics[height=3ex]{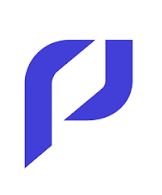}}\,Pythia-410M}
 & base  & 0.35 & 0.14 & 0.06 & 0.03 & 0.02 \\
 & final & 0.62 & 0.54 & 0.58 & 0.72 & 0.98 \\
 & $R_k$ & \textbf{1.76} & \textbf{3.86} & \textbf{9.06} & \textbf{21.3} & \textbf{47.1} \\
\midrule
\multirow{3}{*}{\raisebox{-0.30\height}{\includegraphics[height=3ex]{figs/logo/pythia.png}}\,Pythia-1B}
 & base  & 0.43 & 0.21 & 0.12 & 0.08 & 0.06 \\
 & final & 0.69 & 0.66 & 0.79 & 1.00 & 1.38 \\
 & $R_k$ & \textbf{1.62} & \textbf{3.24} & \textbf{6.82} & \textbf{13.1} & \textbf{23.6} \\
\midrule
\multirow{3}{*}{\raisebox{-0.30\height}{\includegraphics[height=3ex]{figs/logo/pythia.png}}\,Pythia-1.4B}
 & base  & 0.35 & 0.14 & 0.06 & 0.03 & 0.02 \\
 & final & 0.58 & 0.52 & 0.60 & 0.82 & 1.23 \\
 & $R_k$ & \textbf{1.66} & \textbf{3.78} & \textbf{9.63} & \textbf{24.7} & \textbf{60.8} \\
\midrule
\multirow{3}{*}{\raisebox{-0.30\height}{\includegraphics[height=3ex]{figs/logo/pythia.png}}\,Pythia-2.8B}
 & base  & 0.30 & 0.11 & 0.05 & 0.03 & 0.02 \\
 & final & 0.55 & 0.48 & 0.51 & 0.62 & 0.81 \\
 & $R_k$ & \textbf{1.82} & \textbf{4.50} & \textbf{11.0} & \textbf{24.9} & \textbf{52.7} \\
\midrule
\multirow{3}{*}{\raisebox{-0.30\height}{\includegraphics[height=3ex]{figs/logo/pythia.png}}\,Pythia-6.9B}
 & base  & 0.30 & 0.11 & 0.05 & 0.03 & 0.02 \\
 & final & 0.54 & 0.54 & 0.66 & 0.99 & 1.40 \\
 & $R_k$ & \textbf{1.80} & \textbf{4.84} & \textbf{12.6} & \textbf{32.2} & \textbf{64.4} \\
\midrule
\multirow{3}{*}{\raisebox{-0.30\height}{\includegraphics[height=2.5ex]{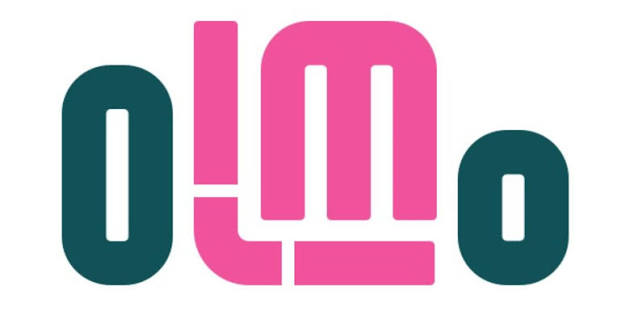}}\,OLMo-1B}
 & base  & 0.39 & 0.19 & 0.12 & 0.08 & 0.07 \\
 & final & 0.80 & 0.74 & 0.79 & 0.93 & 1.08 \\
 & $R_k$ & \textbf{2.05} & \textbf{3.87} & \textbf{6.86} & \textbf{11.4} & \textbf{16.5} \\
\midrule
\multirow{3}{*}{\raisebox{-0.30\height}{\includegraphics[height=3ex]{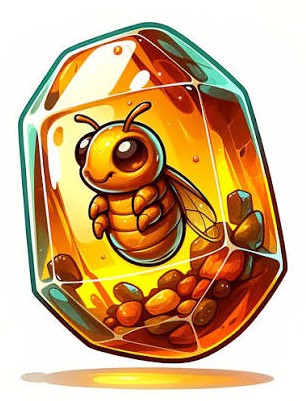}}\,Amber-7B}
 & base  & 0.30 & 0.12 & 0.06 & 0.04 & 0.03 \\
 & final & 0.52 & 0.36 & 0.32 & 0.32 & 0.37 \\
 & $R_k$ & \textbf{1.71} & \textbf{2.98} & \textbf{5.26} & \textbf{8.92} & \textbf{14.1} \\
\bottomrule
\end{tabular}
\caption{Write-relative connected component $\widehat{U}_T$
(Eq.~\ref{eq:normalized-connected}) at the baseline checkpoint (base) and the final released
pretraining checkpoint (final),
with the ratio
$R_k=\widehat{U}^{\,\mathrm{final}}_T/\widehat{U}^{\,\mathrm{base}}_T$
for $k=2$--$6$ (geometric mean over windows). At the baseline
$\widehat{U}_T$ decays by roughly half per order; training removes this
decay, keeping it $O(1)$ at every order.}
\label{tab:coordination-ratio}
\end{table}

We test whether pretraining organizes the nominal depth supplied by the
architecture into interdependent computation. We compare the normalized
connected component of the final checkpoint to that of the same
architecture at baseline checkpoint, and report the coordination ratio $R_k$
at orders $k=2$--$6$ over all adjacent windows in depth.

\paragraph{Write-relative coordination strengthens over baseline.}
As Table~\ref{tab:coordination-ratio} shows, $R_k > 1$ in all 35 model--order cells. The smallest observed value is $R_2 = 1.62$ for Pythia-1B, and even this weakest case lies above its baseline level. Because $\widehat{U}_T$ expresses the connected component relative to the actual write of the final participating layer, $R_k$ compares the strength of the connected interaction against a local write scale rather than using the absolute magnitude of $U_T$. Across every measured model and interaction order, the connected component becomes larger relative to this scale by the final checkpoint.

\begin{figure}[ht]
    \centering
    \includegraphics[width=0.95\linewidth]{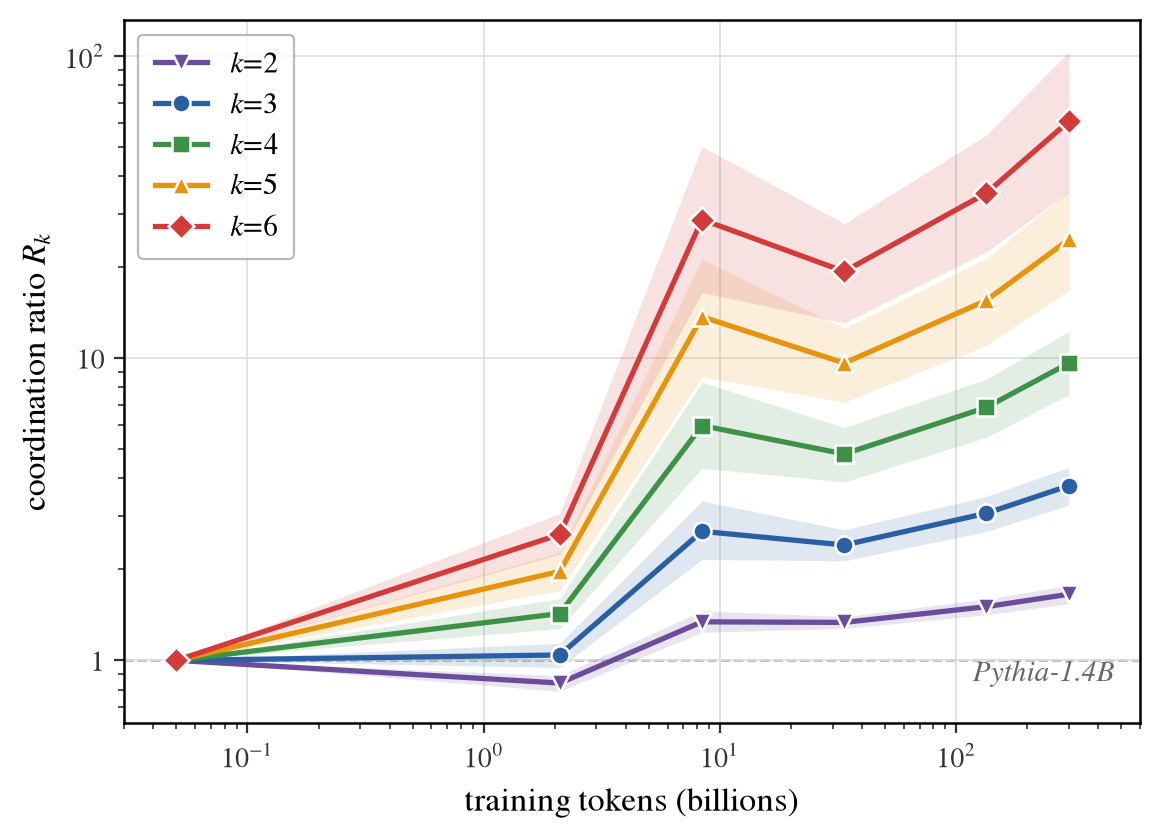}
    \caption{Coordination ratio $R_k$ over pretraining (Pythia-1.4B), averaged over all adjacent windows of order $k$. All orders start near $R_k=1$ and separate upward as training proceeds. Bands are 95\% bootstrap confidence intervals over adjacent windows.}
    \label{fig:coordination-development}
\end{figure}

\paragraph{When coordination forms.}
Figure~\ref{fig:coordination-development} shows, for Pythia-1.4B, how
these values are reached over the course of training. At the earliest
checkpoints $R_k$ stays near 1 at every order: components requiring the
joint participation of several layers exist algebraically even at
random weights, but their size does not exceed what each layer's write
magnitude already predicts. The ratios leave this level only after
2.1B tokens of pretraining, beyond which the bootstrap intervals over
windows no longer contain $R_k = 1$. Cross-layer coordination is not a
byproduct of the weights drifting away from initialization, but a
structure organized at a specific phase of pretraining.

\begin{table*}[thb]
\centering
\small
\setlength{\tabcolsep}{5pt}
\begin{tabular}{@{}l c cc cc cc cc cc@{}}
\toprule
\multirow{2}{*}{Model} & PPL & \multicolumn{2}{c}{$k{=}2$} & \multicolumn{2}{c}{$k{=}3$} & \multicolumn{2}{c}{$k{=}4$} & \multicolumn{2}{c}{$k{=}5$} & \multicolumn{2}{c}{$k{=}6$} \\
\cmidrule(lr){3-4}\cmidrule(lr){5-6}\cmidrule(lr){7-8}\cmidrule(lr){9-10}\cmidrule(lr){11-12}
 & (real) & $U$-del. & rand. & $U$-del. & rand. & $U$-del. & rand. & $U$-del. & rand. & $U$-del. & rand. \\
\midrule
\raisebox{-0.30\height}{\includegraphics[height=3ex]{figs/logo/pythia.png}}\,Pythia-410M & 34.7 & $11.8\times$ & $3.5\times$ & $8.2\times$ & $2.8\times$ & $4.7\times$ & $2.2\times$ & $6.4\times$ & $5.9\times$ & $19.8\times$ & $9.4\times$ \\
\raisebox{-0.30\height}{\includegraphics[height=3ex]{figs/logo/pythia.png}}\,Pythia-1B   & 30.4 & $14.5\times$ & $2.1\times$ & $5.7\times$ & $3.3\times$ & $2.9\times$ & $1.9\times$ & $7.2\times$ & $3.2\times$ & $52.9\times$ & $11.9\times$ \\
\raisebox{-0.30\height}{\includegraphics[height=3ex]{figs/logo/pythia.png}}\,Pythia-1.4B & 26.1 & $10.8\times$ & $2.3\times$ & $52.5\times$ & $3.1\times$ & $12.5\times$ & $4.3\times$ & $35.1\times$ & $2.5\times$ & $224\times$ & $19.8\times$ \\
\raisebox{-0.30\height}{\includegraphics[height=3ex]{figs/logo/pythia.png}}\,Pythia-2.8B & 22.0 & $6.7\times$ & $3.5\times$ & $21.6\times$ & $1.24\times$ & $15.0\times$ & $3.2\times$ & $167\times$ & $3.9\times$ & $111\times$ & $5.3\times$ \\
\raisebox{-0.30\height}{\includegraphics[height=3ex]{figs/logo/pythia.png}}\,Pythia-6.9B & 19.3 & $6.6\times$ & $3.0\times$ & $809\times$ & $2.0\times$ & $63.0\times$ & $5.4\times$ & $1472\times$ & $162\times$ & $1253\times$ & $898\times$ \\
\raisebox{-0.30\height}{\includegraphics[height=2.5ex]{figs/logo/olmo.jpeg}}\,OLMo-1B    & 24.4 & $218\times$ & $18.6\times$ & $2.5\times$ & $1.9\times$ & $4.0\times$ & $2.1\times$ & $3.9\times$ & $3.6\times$ & $19.8\times$ & $7.8\times$ \\
\raisebox{-0.30\height}{\includegraphics[height=3ex]{figs/logo/amber.jpeg}}\,Amber-7B    & 11.0 & $2.3\times$ & $1.20\times$ & $1.8\times$ & $1.05\times$ & $1.1\times$ & $1.04\times$ & $2.7\times$ & $1.09\times$ & $1.8\times$ & $1.20\times$ \\
\bottomrule
\end{tabular}
\caption{Perplexity increase (relative to the unintervened model) from deleting the connected
component $U_T$ versus a norm-matched random direction at the same site, for $k=2$--$6$.
The two deletions have identical magnitude at the intervention point, so the gap is attributable
to the deleted direction being connected. $U$-deletion is more damaging in all 35 cells.}
\label{tab:u-deletion-wide}
\end{table*}

\subsection{Connected Coordination is Behaviorally Consequential}
\label{sec:coord-causal}

The coordination ratio establishes that coordination exists and
strengthens with training, but it does not say whether this component
matters for the model's actual output. If the connected component were
a mere algebraic remainder, removing it should leave model behavior
nearly unchanged. At each order $k$, we compare the perplexity increase
from deleting the connected component $U_T$ from the forward pass with
the increase from deleting a norm-matched random direction at the same
site. Since $U_T$ is by construction a component of the last
participating layer's write, the deletion is applied to the residual
stream immediately after that layer's output. The two deletions have
identical magnitude at the deletion site, so any difference in the
resulting factors comes from whether the deleted direction is the
connected component.

Table~\ref{tab:u-deletion-wide} presents this comparison across orders
$k=2$--$6$. In all $35$ measured cells, deleting the connected
component causes a larger perplexity increase than the random deletion.
The difference is often dramatic: deleting the $k=3$ connected
component of Pythia-6.9B increases perplexity by a factor of $809$,
whereas deleting a random direction of the same magnitude increases it
by only $2.0\times$. Were the connected component a negligible fine
correction, the two factors would be comparable; instead, the factor on
the connected side is reliably larger. The size of the gap varies
across models---the margins are narrow at every order for
Amber-7B in particular---yet the sign is consistent without exception.
This observation shows that the computation multiple layers produce
only together is causally substantive for model output.

\begin{table*}[thb]
\centering
\small 
\setlength{\tabcolsep}{4pt}
\begin{tabular}{llcccccccc}
\toprule
\multirow{2}{*}{Model} & \multirow{2}{*}{$k$} & \multirow{2}{*}{Dense$\downarrow$}
 & \multirow{2}{*}{$\rho_{\mathrm{sim}}$}
 & \multicolumn{3}{c}{Coordination} & \multicolumn{3}{c}{CKA} \\
\cmidrule(lr){5-7}\cmidrule(lr){8-10}
 & & & & Low (safe)$\downarrow$ & High (risky)$\downarrow$ & $\rho$ & High (safe)$\downarrow$ & Low (risky)$\downarrow$ & $\rho$ \\
\midrule
\multirow{3}{*}{\raisebox{-0.30\height}{\includegraphics[height=3ex]{figs/logo/pythia.png}}\,pythia-410M} & 2 & \multirow{3}{*}{34.7} & \multirow{3}{*}{$-0.52$} & \underline{70.2\,{\scriptsize[14--15]}} & 85.5\,{\scriptsize[21--22]} & $+0.63$ & \textbf{68.2\,{\scriptsize[10--11]}} & 139.4\,{\scriptsize[1--2]} & $-0.57$ \\
 & 3 & & & \underline{84.0\,{\scriptsize[10--12]}} & 2130.0\,{\scriptsize[5--7]} & $+0.83$ & \textbf{80.0\,{\scriptsize[9--11]}} & 1486.3\,{\scriptsize[3--5]} & $-0.70$ \\
 & 4 & & & \underline{155.6\,{\scriptsize[18--21]}} & 15978.1\,{\scriptsize[5--8]} & $+0.84$ & \textbf{126.5\,{\scriptsize[12--15]}} & 23693.8\,{\scriptsize[3--6]} & $-0.71$ \\
\midrule
\multirow{3}{*}{\raisebox{-0.30\height}{\includegraphics[height=3ex]{figs/logo/pythia.png}}\,pythia-1B} & 2 & \multirow{3}{*}{30.4} & \multirow{3}{*}{$-0.66$} & 233.1\,{\scriptsize[1--2]} & \underline{90.6\,{\scriptsize[13--14]}} & $+0.05$ & \textbf{76.3\,{\scriptsize[8--9]}} & 233.1\,{\scriptsize[1--2]} & $-0.81$ \\
 & 3 & & & \textbf{77.8\,{\scriptsize[6--8]}} & 1026.2\,{\scriptsize[3--5]} & $+0.83$ & \underline{106.1\,{\scriptsize[7--9]}} & 156.5\,{\scriptsize[11--13]} & $-0.68$ \\
 & 4 & & & \textbf{130.8\,{\scriptsize[7--10]}} & 5535.6\,{\scriptsize[2--5]} & $+0.94$ & \textbf{130.8\,{\scriptsize[7--10]}} & 296.9\,{\scriptsize[9--12]} & $-0.59$ \\

 \midrule
\multirow{3}{*}{\raisebox{-0.30\height}{\includegraphics[height=3ex]{figs/logo/pythia.png}}\,pythia-1.4B} & 2 & \multirow{3}{*}{26.1} & \multirow{3}{*}{$-0.39$} & \textbf{50.1\,{\scriptsize[16--17]}} & 169.2\,{\scriptsize[3--4]} & $+0.30$ & \underline{50.8\,{\scriptsize[20--21]}} & 216.0\,{\scriptsize[1--2]} & $-0.54$ \\
 & 3 & & & \underline{70.1\,{\scriptsize[17--19]}} & 334.6\,{\scriptsize[3--5]} & $+0.55$ & \textbf{69.3\,{\scriptsize[19--21]}} & 1813.6\,{\scriptsize[1--3]} & $-0.58$ \\
 & 4 & & & \underline{137.4\,{\scriptsize[16--19]}} & 7915.1\,{\scriptsize[1--4]} & $+0.58$ & \textbf{96.8\,{\scriptsize[19--22]}} & 1859.1\,{\scriptsize[3--6]} & $-0.66$ \\
\midrule
\multirow{3}{*}{\raisebox{-0.30\height}{\includegraphics[height=3ex]{figs/logo/pythia.png}}\,pythia-2.8B} & 2 & \multirow{3}{*}{22.0} & \multirow{3}{*}{$-0.54$} & \underline{38.4\,{\scriptsize[16--17]}} & 39.4\,{\scriptsize[29--30]} & $+0.54$ & \textbf{36.4\,{\scriptsize[10--11]}} & 141.7\,{\scriptsize[1--2]} & $-0.60$ \\
 & 3 & & & \underline{41.7\,{\scriptsize[14--16]}} & 79.0\,{\scriptsize[18--20]} & $+0.74$ & \textbf{38.9\,{\scriptsize[10--12]}} & 324.0\,{\scriptsize[1--3]} & $-0.53$ \\
 & 4 & & & \underline{50.8\,{\scriptsize[13--16]}} & 1119.0\,{\scriptsize[2--5]} & $+0.82$ & \textbf{47.7\,{\scriptsize[10--13]}} & 1782.5\,{\scriptsize[1--4]} & $-0.44$ \\
\midrule
\multirow{3}{*}{\raisebox{-0.30\height}{\includegraphics[height=3ex]{figs/logo/pythia.png}}\,pythia-6.9B} & 2 & \multirow{3}{*}{19.3} & \multirow{3}{*}{$-0.19$} & \underline{35.8\,{\scriptsize[29--30]}} & 178.5\,{\scriptsize[2--3]} & $+0.11$ & \textbf{32.4\,{\scriptsize[27--28]}} & 178.5\,{\scriptsize[2--3]} & $-0.70$ \\
 & 3 & & & \underline{35.6\,{\scriptsize[23--25]}} & 117.8\,{\scriptsize[15--17]} & $+0.47$ & \textbf{34.7\,{\scriptsize[27--29]}} & 4858.9\,{\scriptsize[1--3]} & $-0.64$ \\
 & 4 & & & \underline{39.7\,{\scriptsize[22--25]}} & 8145.7\,{\scriptsize[3--6]} & $+0.61$ & \textbf{38.8\,{\scriptsize[26--29]}} & 785.4\,{\scriptsize[1--4]} & $-0.59$ \\
\midrule
\multirow{3}{*}{\raisebox{-0.30\height}{\includegraphics[height=2.5ex]{figs/logo/olmo.jpeg}}\,OLMo-1B} & 2 & \multirow{3}{*}{24.4} & \multirow{3}{*}{$-0.48$} & \textbf{49.3\,{\scriptsize[7--8]}} & 145.8\,{\scriptsize[13--14]} & $+0.55$ & \underline{88.9\,{\scriptsize[10--11]}} & 901.9\,{\scriptsize[1--2]} & $-0.69$ \\
 & 3 & & & \underline{138.2\,{\scriptsize[7--9]}} & \textbf{124.0\,{\scriptsize[10--12]}} & $+0.40$ & \textbf{124.0\,{\scriptsize[10--12]}} & 1866.0\,{\scriptsize[1--3]} & $-0.70$ \\
 & 4 & & & \textbf{173.8\,{\scriptsize[7--10]}} & 2866.1\,{\scriptsize[1--4]} & $+0.75$ & \underline{183.4\,{\scriptsize[6--9]}} & 2866.1\,{\scriptsize[1--4]} & $-0.71$ \\
\midrule
\multirow{3}{*}{\raisebox{-0.30\height}{\includegraphics[height=3ex]{figs/logo/amber.jpeg}}\,Amber-7B} & 2 & \multirow{3}{*}{11.0} & \multirow{3}{*}{$-0.48$} & \textbf{23.8\,{\scriptsize[4--5]}} & \underline{26.7\,{\scriptsize[29--30]}} & $+0.57$ & 55.7\,{\scriptsize[27--28]} & 26.9\,{\scriptsize[1--2]} & $+0.18$ \\
 & 3 & & & 33.3\,{\scriptsize[3--5]} & 34.3\,{\scriptsize[28--30]} & $+0.49$ & \underline{32.1\,{\scriptsize[27--29]}} & \textbf{25.7\,{\scriptsize[8--10]}} & $+0.01$ \\
 & 4 & & & \underline{40.7\,{\scriptsize[3--6]}} & 44.9\,{\scriptsize[27--30]} & $+0.32$ & 49.9\,{\scriptsize[26--29]} & \textbf{34.8\,{\scriptsize[7--10]}} & $-0.07$ \\
\bottomrule
\end{tabular}
\caption{Perplexity after removing $k$ consecutive layers, with blocks selected by coordination
versus CKA. Bracketed indices denote the removed block; \textbf{best} and \underline{second best}
per row. $\rho_{\mathrm{sim}}$ is the model-level Spearman correlation between pairwise CKA and
$\hat{U}_{\{i,j\}}$ over all layer pairs. $\rho$ is the Spearman correlation between a metric's
block score and post-removal perplexity; correct ranking gives $\rho>0$ for coordination and
$\rho<0$ for CKA.}
\label{tab:coord-cka}
\end{table*}

\subsection{Representation Similarity Tracks, but Does Not Determine, Coordination}

\label{sec:cka}

Layer redundancy has often been diagnosed through representation
similarity: adjacent layers with similar representations have been
interpreted as doing similar work \cite{nguyen2021wide}, motivating
layer skipping and deep-layer pruning \cite{gromov2024unreasonable, sun2025painters}.
We compare the CKA \cite{kornblith2019similarity} of each layer pair
with the normalized connected component
$\widehat{U}_{\{i,j\}}$ of the same pair.
As shown in Table~\ref{tab:coord-cka}, the two quantities are
negatively correlated in all seven models
($\rho_{\mathrm{sim}}$ between $-0.66$ and $-0.19$):
layer pairs with more similar representations tend to exhibit weaker
pairwise connected computation.

This association provides a functional interpretation of one pattern
underlying representation similarity, but it does not make the two
notions equivalent.
High similarity need not imply that two layers repeat the same
computation, and weak coordination does not by itself imply that their
individual writes are dispensable.
Rather, our results show that preservation of representation geometry
is systematically associated with weaker pairwise interaction.
Coordination therefore captures how one layer's computation depends on
another, providing information complementary to state-level similarity.

Table~\ref{tab:coord-cka} further compares the two signals through
actual block removal.
For each model, we select $k$ consecutive layers according to the
coordination and CKA criteria, remove them, and measure perplexity.
Across the Pythia family, the two criteria often give similar
recommendations: the perplexities of the low-coordination and
high-CKA blocks agree within 8\% in 10 of the 15 cells, and the
low-CKA blocks fall in the early layers in 13 cells, frequently
overlapping with high-coordination regions.
At $k=2$ on Pythia-6.9B and $k=4$ on OLMo, the two criteria select
the same block.
The agreement is not uniform, however.
On Amber-7B, CKA's block-ranking correlation becomes near zero or
changes sign ($+0.18$, $+0.01$, $-0.07$), whereas coordination retains
a positive association with removal damage
($+0.57$, $+0.49$, $+0.32$).
The blocks favored by high CKA yield perplexities of
55.7, 32.1, and 49.9, compared with 26.9, 25.7, and 34.8 for the
corresponding low-CKA blocks.
Thus, even though $\rho_{\mathrm{sim}}=-0.48$ on Amber-7B, pairwise
state similarity alone does not reliably rank block-removal cost.

The distinction is structural.
Similarity is fundamentally pairwise, whereas
Section~\ref{sec:coord-learned} shows that connected computation also
appears at orders of three or more layers.
A layer with weak pairwise coordination can still participate in a
stronger higher-order interaction that pairwise similarity cannot
observe.
Conversely, weak connected interaction alone does not guarantee
inexpensive removal, because a layer may have a large individual
contribution even when its interactions with other layers are small.
The mixed behavior of similarity-based pruning reported in prior work
\cite{freiberger2024layershuffle, sun2025painters} is consistent with
this distinction.
Notably, Pythia-6.9B, which has the weakest pairwise association
($\rho_{\mathrm{sim}}=-0.19$), also has the largest $k=3$ coordination
ratio among the measured models
($R_3=4.84$, Table~\ref{tab:coordination-ratio}).

In sum, representation similarity is a useful but incomplete signal of
cross-layer interaction.
High similarity is consistently associated with weaker pairwise
coordination in our models, but neither pairwise similarity nor weak
coordination alone guarantees safe layer removal.
Higher-order connected components provide complementary information
about cross-layer dependencies that pairwise state comparisons cannot
capture.

\section{Discussion}
\label{sec:discussion}

That cross-layer coordination forms through training and contributes
substantively to model output shows that deep models do not merely
accumulate independent per-layer refinements. It does not follow,
however, that coordination is a direct cause of improved performance.
At least two directions of explanation are possible.

One possibility is that training organizes layers into interdependent
computation, and the resulting coordination enables the model to
express more complex functions. On this view, coordination is the
mechanism that converts the nominal depth supplied by the architecture
into actual computational capacity: two models with the same number of
layers may differ in effective computational depth, and in performance,
depending on how strongly their layers form connected computation.

The opposite direction is also possible. In the course of solving hard
prediction problems the model may come to require several sequential
intermediate computations, with cross-layer coordination emerging as a
result. If information must be extracted, transformed in context, and
integrated with other information, each step naturally depends on the
result of the preceding one. On this view, coordination is less an
independent cause of performance than a structural signature that the
model has acquired multi-step computation.

The two explanations need not be exclusive. The computational demands
of the task may first induce coordination, and the coordination thus
formed may in turn make complex computation reliably executable. Our
results show that coordination is used in actual model behavior, but
they do not distinguish whether coordination precedes the emergence of
a capability or forms as a consequence of learning it. Settling this
would require manipulating the sequential complexity of the task, or
jointly tracking when specific capabilities and coordination emerge
during training.

\paragraph{Limitations.}
Our measurements focus on windows of layers adjacent in depth, which keeps the intervention site well defined but leaves coordination
between distant layers largely unmeasured; identity ablation also induces counterfactual states within each window.
Our normalization is relative to the actual write of the final participating layer and does not separately normalize preceding-layer write magnitudes.
Finally, computing $U_T$ exactly requires $2^{k}$ subset forward passes, making higher interaction orders beyond $k \le 6$ increasingly expensive and likely to require approximation.

\section{Conclusion}
We introduced an inclusion--exclusion decomposition that isolates the
computation arising only when multiple layers act together, and traced
its development across public checkpoints of Pythia, OLMo, and Amber.
In our experimental setting, we find that write-relative cross-layer
coordination strengthens over pretraining, that deleting the resulting
connected component is consistently more damaging than deleting a
norm-matched random direction at the same site, and that representation
similarity provides a complementary, but not interchangeable, signal of
cross-layer dependency and block-removal cost. These observations are
consistent with the view that architectural depth provides the
opportunity for sequential computation, while pretraining progressively
organizes cross-layer dependencies within that depth. We hope that
treating functional depth as a learned, empirical property offers a
new lens for analyzing, training, and deploying deep models.

\bibliography{aaai2027}

\end{document}